\documentclass[lettersize,journal,twoside]{IEEEtran}
\usepackage{amsmath,amsfonts}
\usepackage{algorithmic}
\usepackage{algorithm}
\usepackage{array}
\usepackage[caption=false,font=normalsize,labelfont=sf,textfont=sf]{subfig}
\usepackage{textcomp}
\usepackage{stfloats}
\usepackage{url}
\usepackage{verbatim}
\usepackage{graphicx}
\usepackage{cite}
\usepackage{xcolor}
\begin{document}

\title{SkillPlug: Unsupervised Skill Mining for Few-Shot Adaptation in Robotic Manipulation}

\author{Zi-han Ding$^{1}$, Ziwei~Wang$^{1*}$%
\thanks{Manuscript received: December 23, 2025; Revised: March 30, 2026; Accepted: May 17, 2026.}%
\thanks{This paper was recommended for publication by Editor Markus Vincze upon evaluation of the Associate Editor and Reviewers comments.}%
\thanks{This work was supported by MoE AcRF Tier 2 (MOE-T2EP50125-0004).}%
\thanks{$^{1}$Zi-han Ding and Ziwei Wang are with the School of Electrical and Electronic Engineering, Nanyang Technological University, Singapore.
{\tt\small \{zihan.ding, ziwei.wang\}@ntu.edu.sg}}%
\thanks{$^{*}$Corresponding author: Ziwei Wang.}%
\thanks{Digital Object Identifier (DOI): see top of this page.}%
}

% The paper headers
\markboth{IEEE ROBOTICS AND AUTOMATION LETTERS. PREPRINT VERSION. ACCEPTED JUNE, 2026}%
{Ding \MakeLowercase{\textit{et al.}}: SkillPlug: Unsupervised Skill Mining for Few-Shot Adaptation in Robotic Manipulation}

\maketitle

\begin{abstract}
Learning transferable visuomotor imitation policies that generalize across diverse manipulation tasks and adapt rapidly to new tasks from only a handful of demonstrations remains challenging. Most modern policies are trained end-to-end to map observations directly to low-level actions, offering little explicit structure for reusing and recombining behaviors across tasks and making transfer data-inefficient under limited supervision. We propose \emph{SkillPlug}, a plug-in framework that augments an existing visuomotor policy with a skill-conditioning module and mines a shared, transferable skill library from raw multi-task demonstrations. SkillPlug learns skills via self-supervised objectives that promote compact, reusable, and non-redundant behavior-level primitives, forming a task-shared prior for compositional control. After skill mining, we keep the learned skills fixed and specialize to unseen tasks by fine-tuning only lightweight router and action head, enabling efficient adaptation without full end-to-end retraining. We evaluate SkillPlug on two simulation benchmarks and on a real robot, and observe that the mined transferable skills consistently improve both multi-task performance and few-shot adaptation. Overall, SkillPlug offers a scalable way to mine reusable skills that improve data-efficient generalization in robotic manipulation.
\end{abstract}


\section{Introduction}
Imitation-based visuomotor policies have driven rapid progress in robotic manipulation.
Trained on large collections of demonstrations~\cite{walke2023bridgedata, o2024open, liu2023libero} and scaled with modern transformer backbones~\cite{brohan2022rt, zitkovich2023rt, kim2025openvla, mees2024octo, black2024pi, intelligence2504pi0, li2025cronusvla}, a single policy can now solve a wide range of tasks. Despite these advances, most policies are still optimized as end-to-end mappings from observations to low-level actions, leaving reusable behavioral structure implicit. As a result, while end-to-end policies can perform well in multi-task settings, they often fail to uncover and reuse cross-task shared behavioral patterns, making it difficult to efficiently transfer to novel tasks with only a few demonstrations.

A natural remedy is to introduce an explicit skill abstraction as an intermediate layer. Existing imitation-learning-based approaches to skill learning broadly fall into two categories. One line relies on human annotation or VLM-based segmentation to define skills or phases (e.g., \emph{approach}, \emph{grasp}, \emph{place}), and then trains hierarchical controllers conditioned on these labels~\cite{xu2025query, lorang2025few, fan2025long, qi2025compose, zhao2025generalizable, fu2024language}. While such methods offer interpretable structure, they typically require extensive manual design for segmentation, skill definition, and phase-boundary specification. A second line learns skills directly from data, e.g., via trajectory clustering, learnable skill embeddings, or mixture-of-experts routing where experts implicitly serve as skills~\cite{wang2025experts, zheng2025universal, jiang2025discrete, liusemantic, yao2025think, mete2024quest, xu2025speci, cheng2025moe}. However, without objectives that explicitly encourage reusability and separation, learned skills may collapse to similar behaviors or entangle with visual context, yielding redundant libraries.

\begin{figure}[t]
  \centering
  \includegraphics[width=\linewidth]{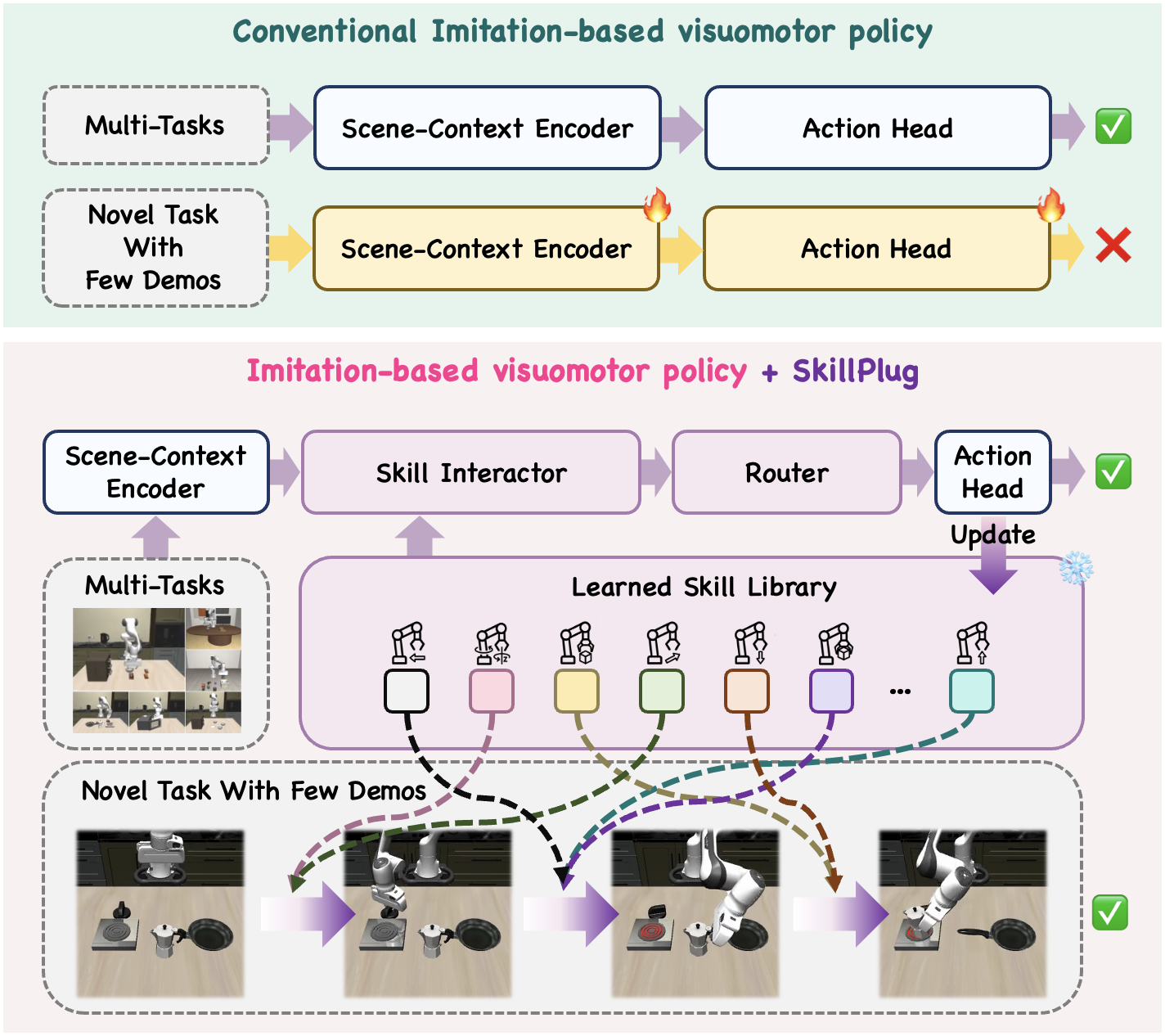}
  \caption{Conventional imitation-based visuomotor policies often struggle to adapt to a new task from only a handful of demonstrations. SkillPlug augments the policy with a skill interactor and router, and mines a transferable skill library from multi-task demonstrations, enabling efficient reuse and recombination of learned skills for successful few-shot transfer.}
  \label{fig:teaser}
\end{figure}

In this work, we introduce \emph{SkillPlug}, an unsupervised skill-mining approach that learns transferable skills from raw multi-task demonstrations, improving both multi-task generalization and data-efficient few-shot adaptation. As illustrated in Fig.~\ref{fig:teaser}, SkillPlug is a plug-in, architecture-agnostic framework that augments existing visuomotor policies with a skill-conditioning module while preserving their original backbones. It learns a reusable and non-redundant skills library through self-supervised objectives, making behavioral structure explicit and shared across tasks. At adaptation time, we keep the learned skills fixed and specialize to a new task by fine-tuning only lightweight components, avoiding full end-to-end retraining. 

% Across two simulation benchmarks and real-robot evaluations, SkillPlug yields consistent relative gains. On DISCOVERSE, ACT+SkillPlug improves multi-task success by approximately 106\% and improves 10-demo few-shot success by approximately 53\%. On LIBERO cross-suite transfer with 5 demos per task, average few-shot success improves by approximately 630\%. On the real robot, the average number of successful trials per task improves by approximately 62\% on seen tasks and by approximately 4\% on held-out tasks.
Across two simulation benchmarks and real-robot evaluations, SkillPlug yields consistent gains. On DISCOVERSE, ACT+SkillPlug improves multi-task and 10-demo few-shot success by +45.1\% and +18.1\% points; on LIBERO, OpenVLA-OFT+SkillPlug improves 5-demo cross-suite transfer by +38.3\% points on average; and on the real robot, the average success rate increases by +28.5\% points on few-shot tasks.

% We validate SkillPlug on both compact and large-scale policies across two simulation benchmarks and a real robot platform, showing that the transferable skills mined by SkillPlug consistently improve both multi-task performance and data-efficient few-shot adaptation. 

In summary, the contributions of this paper are as follows:
\begin{itemize}
    \item We show that unsupervisedly mined transferable skills provide a shared behavioral prior that consistently improves both multi-task generalization and data-efficient few-shot adaptation.
    \item We propose SkillPlug, a plug-in, architecture-agnostic framework that augments existing visuomotor policies with a lightweight skill-conditioning module and mines reusable, non-redundant skills from raw multi-task demonstrations using self-supervised objectives.
    \item We show consistent gains over both compact and large-scale baselines on two simulation benchmarks and real-robot evaluations, improving both multi-task performance and few-shot adaptation.
\end{itemize}

\section{Related Work}

\subsection{Visuomotor Imitation Learning}
Imitation-based visuomotor policies have been extensively studied for multi-task and few-shot robotic manipulation, ranging from conventional imitation-learning policies to recent vision-language-action models (VLAs). A large body of work trains a single behavior-cloned \cite{zhaolearning, chi2025diffusion, mandi2022towards} that solves multiple manipulation tasks or adapts from few demonstrations. In parallel, large VLA models such as RT-1/2\cite{brohan2022rt, zitkovich2023rt}, OpenVLA\cite{kim2025openvla}, Octo\cite{mees2024octo}, RDT-1B\cite{liurdt}, and $\pi_0/\pi_{0.5}$\cite{black2024pi, intelligence2504pi0} scale transformer-based policies on massive robot datasets\cite{walke2023bridgedata, o2024open, liu2023libero}, often built on Large Vision-and-Language backbones\cite{touvron2023llama, bai2023qwenvl, beyer2024paligemma}. Despite their strong multi-task performance, these policies are still trained as end-to-end mappings from observations and task descriptions to low-level actions, leaving skill-level structure implicit. As a result, they often fail to uncover cross-task shared behavioral patterns that could be reused and recombined, which in turn limits efficient transfer to genuinely novel tasks from only a handful of demonstrations. In contrast, our work introduces an explicit skill-level representation, aiming to preserve the benefits of multi-task imitation learning while enabling reusable, compositional behaviors and improving few-shot generalization.

% Imitation learning has enabled multi-task visuomotor manipulation via behavior cloning, where a single policy is trained to solve multiple tasks or adapt from a few demonstrations~\cite{zhaolearning, chi2025diffusion, mandi2022towards, duan2017one}. Recent vision--language--action (VLA) models further scale this paradigm with large transformer backbones \cite{touvron2023llama, bai2023qwenvl, beyer2024paligemma} and multi-robot datasets~\cite{brohan2022rt, zitkovich2023rt, kim2025openvla, mees2024octo, liurdt, black2024pi, intelligence2504pi0, walke2023bridgedata, o2024open, liu2023libero}. Despite strong multi-task performance, most methods remain end-to-end mappings from observations and task descriptions to low-level actions, leaving reusable skill structure implicit and making transfer under limited data challenging.

\subsection{Skill Learning in Imitation-Based Robotic Manipulation}

Existing imitation-learning-based skill methods for manipulation largely follow two trends. A first line explicitly defines skills via human annotation \cite{xu2025query,lorang2025few,fan2025long, qi2025compose} or VLM-based segmentation and labeling \cite{zhao2025generalizable, fu2024language}, e.g., decomposing demonstrations into phases such as \emph{approach}, \emph{grasp}, and \emph{place}, and then building hierarchical policies on top of these labels. While this yields interpretable skill libraries, it requires substantial manual trajectory segmentation and label design, and the resulting skill granularity may still not match what the policy actually needs for control. A second line learns skills directly from data, for example by clustering trajectories \cite{wang2025experts}, introducing learnable skill embeddings \cite{zheng2025universal, jiang2025discrete, liusemantic, yao2025think, mete2024quest, xu2025speci}, or using mixture-of-experts routers where each expert implicitly plays the role of a skill \cite{cheng2025moe}. However, the learned skills are still driven primarily by low-level action supervision, with little explicit structure imposed on the skill space, and remain strongly tied to visual appearance, which limits their transfer across tasks and scenes. In contrast, SkillPlug performs fully unsupervised skill mining from raw demonstrations by learning a reusable, task-shared skill library and an internal router within an existing policy, with objectives that explicitly encourage scene-agnostic, behaviorally distinct, and composable skills that transfer across tasks.

\section{Methodology}

\begin{figure*}[t]
  \centering
  \includegraphics[width=\linewidth]{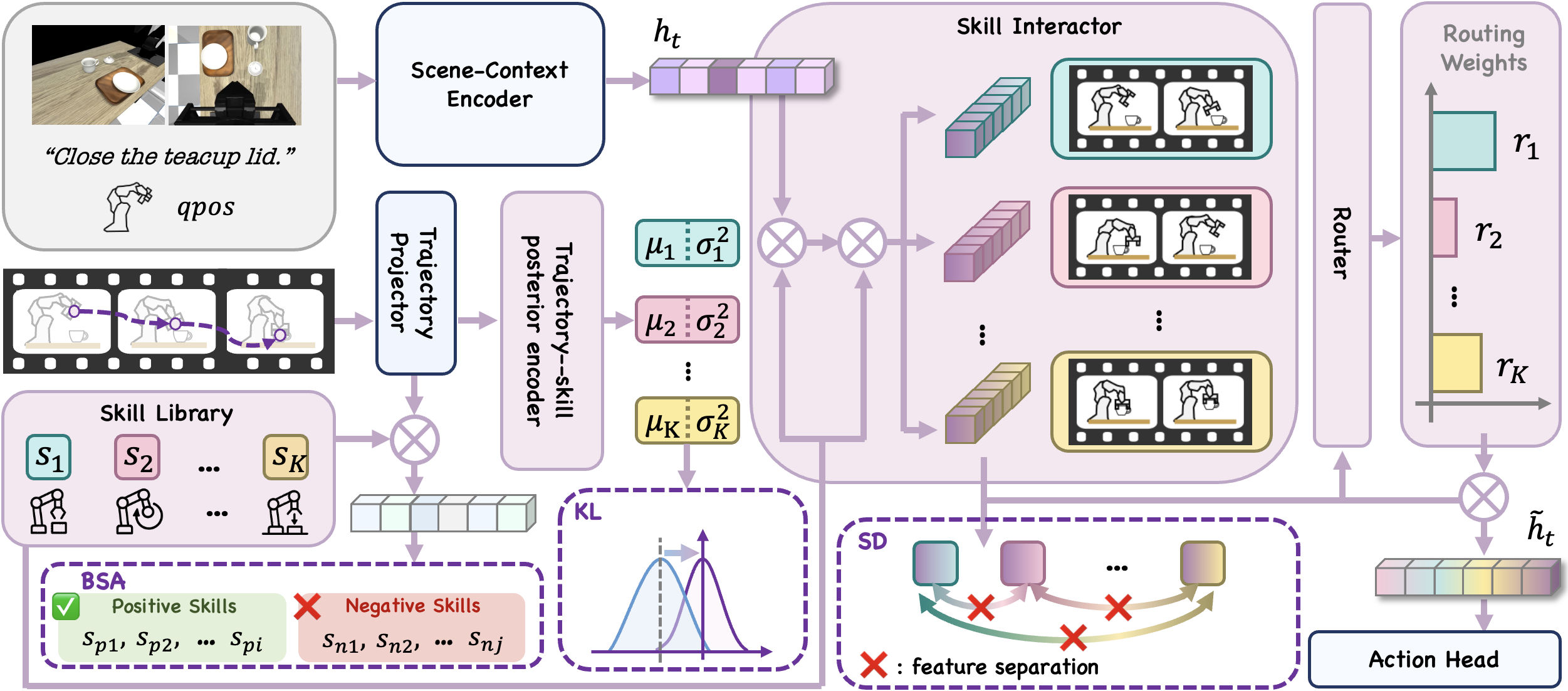}
  \caption{SkillPlug architecture and training. SkillPlug augments a base visuomotor policy with a skill library, a skill interactor, and a router that composes skill-conditioned features for action prediction. During training, a Trajectory--skill posterior encoder is introduced to mine skills from raw demonstrations, and is removed at inference. SkillPlug is learned with four self-supervised objectives: a reconstruction loss for faithful imitation, a KL regularizer for compactness, a behavioral skill alignment loss to tie skills to trajectory-level behavior, and a skill disentanglement loss to reduce redundancy.}
  \label{fig:overall_framework}
\end{figure*}

In this section, we first formalize our training-and-adaptation regime, then introduce the SkillPlug module that can be plugged into existing visuomotor policy as Fig.~\ref{fig:overall_framework}, and finally describe the self-supervised objectives used to mine a transferable skill library.

\subsection{Problem Setup and Notation}

We study offline multi-task visuomotor imitation learning. At each decision step $t$, the robot receives an observation $o_t \in \mathcal{O}$ and predicts a length-$T$ action chunk
\begin{equation}
    \tau_t = (a_{t,1}, a_{t,2}, \dots, a_{t,T}),
\end{equation}
where $a_{t,i}\in\mathbb{R}^{d_a}$ denotes the $i$-th action in the predicted chunk. Each training sample is associated with a task label $l \in \mathcal{L}$. 
We assume access to an offline dataset of multi-task demonstrations, from which we sample chunk-level training tuples $\mathcal{D}=\{(o_t,\tau_t,l)\}$ over a set of training tasks.

% \textbf{Multi-task training.}
% SkillPlug introduces a shared set of $K$ skills intended to capture reusable behavioral primitives across tasks.
% During multi-task training, we mine these transferable skills from $\mathcal{D}_{\text{multi}}$ in a self-supervised manner and learn a skill-conditioned policy that composes them to solve diverse training tasks.
\textbf{Multi-task training.}
We consider a multi-task training stage where a policy is trained jointly on demonstrations from multiple tasks. The goal is to mine a set of transferable skills in a self-supervised manner, and to learn a skill-conditioned policy that can compose these skills to solve the diverse training tasks.

\textbf{Few-shot adaptation.}
In the downstream adaptation setting, we consider novel tasks not seen during training, each provided with only a few demonstrations. We reuse the mined skills as a transferable prior and adapt to each new task by fine-tuning only lightweight components, enabling data-efficient specialization without full end-to-end retraining.

\subsection{Skill-Conditioned Policy Architecture}
\label{subsec:skill_arch}

SkillPlug treats any base policy as a composition of a scene-context encoder $f_{\text{enc}}$ and an action head $f_{\text{act}}$:
\begin{equation}
    h_t = f_{\text{enc}}(o_t), \qquad
    \hat{a}_t = f_{\text{act}}(\tilde{h}_t),
\end{equation}
where $h_t$ is the hidden feature and $\tilde{h}_t$ is the input to the action head; the original policy uses $\tilde{h}_t = h_t$. SkillPlug instead applies a skill-conditioned transformation to $h_t$ while leaving $f_{\text{enc}}$ and $f_{\text{act}}$ unchanged, requiring only their input--output tensor shapes and imposing no architectural constraints.

% \textbf{Skill library.} We maintain a shared library of $K$ learnable skill embeddings $\{s_k \in \mathbb{R}^{d_s}\}_{k=1}^{K}$.
% These embeddings are shared across tasks and environments and are intended to encode scene-agnostic behavior-level primitives.
\textbf{Skill library.}
We maintain a shared library of $K$ learnable skill embeddings $\{s_k \in \mathbb{R}^{d_s}\}_{k=1}^{K}$, where each $s_k$ is a continuous latent vector representing a skill. Each embedding serves as a skill code that conditions the policy to execute a specific primitive behavior, providing an explicit bridge between perception and low-level actions. Crucially, the same set of skill embeddings is shared across tasks and reused in both multi-task training and few-shot adaptation. This shared skill set serves as a transferable prior, allowing us to adapt to new tasks by keeping the skills fixed and fine-tuning only lightweight components for data-efficient specialization with cross-task reuse.

\textbf{Trajectory-skill posterior encoder.}
To bias skills toward behavioral structure rather than scene appearance, we introduce a VAE-style posterior encoder $q_{\phi}$ that takes only an action segment and skill embedding as input. Given a trajectory segment $\tau$, we extract a trajectory feature $f_{\tau}=f_{\phi}(\tau)$ and combine it with the $k$-th skill embedding $s_k$ to form a variational posterior over a skill-specific latent
\begin{equation}
    z_k \sim q_{\phi}(z_k \mid f_{\tau}, s_k).
\end{equation}
Here, $f_\phi$ is an MLP trajectory projector that maps the action chunk
$\tau_t \in \mathbb{R}^{T \times d_a}$ to a chunk-level feature
$f_\tau$, using only actions as input.
This posterior encoder serves as an information bottleneck on the trajectory by constraining the posterior distribution and thereby limiting how much trajectory-specific detail each skill can absorb. This encourages the shared skills to capture stable, reusable trajectory-aligned behavior primitives. Importantly, the encoder is used only during multi-task training and is not designed to enforce a one-to-one correspondence between skills and object-aware semantic functions. Instead, scene-dependent functional disambiguation is handled at router. After training, we discard the posterior branch and act directly with the learned deterministic skill embeddings.

\textbf{Skill interactor and router.}
Given the extracted scene feature $h_t$ and the skill library $\{s_k\}_{k=1}^{K}$, SkillPlug computes a per-skill feature
\begin{equation}
    u_{t,k} = \text{Interactor}(h_t, s_k), \qquad k = 1,\ldots,K,
\end{equation}
where $u_{t,k}$ lives in the same space as $h_t$.
The interactor is a lightweight skill-conditioning module that modulates the scene representation with each skill, implemented via cross-attention to produce skill-specific features. 

A router predicts chunk-level routing weights $r_\tau \in \mathbb{R}^{K}$ over the $K$ skills, where $r_{\tau,k}$ denotes the contribution of skill $k$ for the action chunk $\tau$.
This routing distribution is shared across all timesteps within the corresponding chunk.
SkillPlug then aggregates the per-skill features $\{u_{t,k}\}$ into a single skill-conditioned representation $\tilde{h}_t$ via a weighted sum.
Since $\tilde{h}_t$ matches the dimensionality of the backbone feature $h_t$, it can be fed into the original action head without architectural changes.

% A router then predicts routing weights $r_t \in \mathbb{R}^{K}$:
% \begin{equation}
%     r_t = \text{Router}\!\left(\{u_{t,k}\}_{k=1}^{K}\right), \qquad
%     r_{t,k} \ge 0,\ \ \sum_{k=1}^{K} r_{t,k} = 1.
% \end{equation}
% Finally, SkillPlug aggregates the per-skill features into a single skill-conditioned representation
% \begin{equation}
%     \tilde{h}_t = \sum_{k=1}^{K} r_{t,k}\,u_{t,k},
% \end{equation}
% which has the same dimensionality as $h_t$ and can be fed into the original action head without architectural changes.

\textbf{Train-then-freeze for few-shot adaptation.}
After multi-task training, we freeze the backbone encoder and the learned skill embeddings, and adapt only router and action head for a new task, so that few-shot learning primarily amounts to learning to reuse and compose the existing skills.

\subsection{Learning Skill Representations}
\label{subsec:training_objectives}

To learn a transferable skill library that supports reuse and recombination, we optimize self-supervised objectives that encourage: (i) faithful imitation, (ii) compact skill representations that mitigate overfitting to scene appearance, (iii) behavioral alignment between skills and trajectories, and (iv) separation to prevent redundant skills.

% \vspace{0.3em}
% \textbf{Reconstruction Loss.}
% We enforce faithful imitation by supervising the predicted action chunk with an $\ell_1$ behavior cloning loss:
% \begin{equation}
% \mathcal{L}_{\text{rec}}
% =
% \mathbb{E}_{(o_t,a_{t:t+n-1})\sim\mathcal{D}}
% \Big[\sum_{j=0}^{n-1}\|\hat{a}_{t+j}-a_{t+j}\|_1\Big].
% \end{equation}
\textbf{Reconstruction Loss.}
To encourage faithful imitation, we supervise the predicted $n$-step action chunk $\hat{a}_{t:t+n-1}$ from observation $o_t$ with an $\ell_1$ behavior cloning loss against the demonstrated actions $a_{t:t+n-1}$ sampled from the dataset $\mathcal{D}$:
\begin{equation}
\label{eq:recon}
\mathcal{L}_{\text{rec}}
=
\mathbb{E}_{(o_t,a_{t:t+n-1})\sim\mathcal{D}}
\Big[\sum_{j=0}^{n-1}\|\hat{a}_{t+j}-a_{t+j}\|_1\Big],
\end{equation}
where $n$ is the prediction horizon and $\|\cdot\|_1$ denotes element-wise absolute deviation.

\textbf{KL regularization.}
To promote compactness and bias skills toward trajectory-level behavior rather than scene appearance, we apply a KL regularizer on the trajectory--skill posterior encoder, which only conditions on trajectory features $f_{\tau}$ and skill embeddings $s_k$, without direct access to observations,
\begin{equation}
\mathcal{L}_{\text{KL}}
=
\mathbb{E}_{\tau\sim\mathcal{D}}
\Big[\frac{1}{K}\sum_{k=1}^{K}
D_{\text{KL}}\big(q_{\phi}(z_k\mid f_{\tau},s_k)\,\|\,p(z)\big)\Big].
\end{equation}
This constraint limits the information capacity of each skill latent, discouraging scene-specific encoding and promoting stable structure that recurs across trajectories.

Behavioral Skill Alignment.
Compactness alone does not guarantee that skills capture meaningful behavior.
From an information-theoretic perspective, we would like skill embeddings to be informative about trajectory-level behavior, which can be expressed as encouraging high mutual dependence between a segment $\tau$ and the selected skill embedding $s$:
\begin{equation}
I(\tau; s)=\mathbb{E}_{p(\tau,s)}\!\left[\log \frac{p(s\mid\tau)}{p(s)}\right].
\label{eq:mi-log-ratio-uncond}
\end{equation}

% In practice, we implement this mutual-dependence goal with an InfoNCE-style contrastive objective. 
In practice, we use a soft-label classification objective over the learned skill library, inspired by the contrastive classification view behind InfoNCE. In particular, Eq.~(7) motivates learning a scoring function that assigns higher compatibility to trajectory--skill pairs that better match the segment-level behavior.
% In particular, the log-ratio in Eq.~(\ref{eq:mi-log-ratio-uncond}) motivates learning a critic that scores matched $(\tau,s)$ pairs higher than mismatched ones.
% We realize this by performing a softmax classification over skills in the library.
We realize this as a softmax classification problem over the skill library, where the router assignment provides a soft target distribution over skills.
We first extract a trajectory feature $f_\tau$ from $\tau$ and project both trajectories and skills into a shared embedding space,
\begin{equation}
v_\tau = g_\tau(f_\tau), \qquad v_k = g_s(s_k),
\end{equation}
and score how well skill $k$ explains the segment using inner-product similarity
\begin{equation}
T_\eta(\tau,s_k)=\mathrm{sim}(v_\tau,v_k)=v_\tau^\top v_k.
\end{equation}
% The router provides a soft assignment $r_{\tau,k}$ over skills for segment $\tau$, and we stop gradients through $r_{\tau,k}$ so it serves as a fixed soft target.
% This yields the BSA loss:
% The router provides a soft assignment over skills for segment $\tau$, and we stop gradients through $r_{\tau,k}$ so it serves as a fixed soft target.This prevents a circular coupling between router assignments and skills during training.
% The router provides a soft assignment over skills for segment $\tau$. 
We stop gradients through $r_{\tau,k}$ so that it serves only as a fixed target for shaping the skill embeddings, avoiding circular co-adaptation during training.
With stop-gradient, $\mathcal{L}_{\mathrm{BSA}}$ shapes the skill library, while the router is trained only through the action reconstruction objective via the fused representation in Eq.~(\ref{eq:recon}).
This yields the BSA loss:
\begin{equation}
    \mathcal{L}_{\mathrm{BSA}}
    =
    \mathbb{E}_{\tau \sim \mathcal{D}}
    \Bigg[
        - \sum_{k=1}^{K}
        \operatorname{sg}\!\big[r_{\tau,k}\big]\,
        \log \sigma_k(v_\tau)
    \Bigg],
\end{equation}
where
\begin{equation}
    \sigma_k(v_\tau)
    =
    \frac{
        \exp\big( T_\eta(\tau,s_k) \big)
    }{
        \sum_{k'=1}^{K}
        \exp\big( T_\eta(\tau,s_{k')} \big)
    }.
\end{equation}
Minimizing $L_{\mathrm{BSA}}$ encourages the learned skill embeddings to align with trajectory-level behavioral structure, which improves cross-task reuse and transferability in practice.
\textbf{Skill Disentanglement.}
Compactness and behavioral alignment alone do not prevent redundant skills. We therefore penalize similarity between skill-conditioned features produced by different skills under the same observation.

We temporally average the interactor outputs
\begin{equation}
u_{\tau,k}=\frac{1}{T}\sum_{t=1}^{T}u_{t,k},
\qquad 
\bar{u}_{\tau,k}=\frac{u_{\tau,k}}{\|u_{\tau,k}\|_2},
\end{equation}
form a $K\times K$ similarity matrix $S_{\tau}[k,j]=\bar{u}_{\tau,k}^{\top}\bar{u}_{\tau,j}$, and define
\begin{equation}
\mathcal{L}_{\text{SD}}
=
\mathbb{E}_{\tau\sim\mathcal{D}}
\Big[\frac{1}{K(K-1)}\sum_{k\ne j} S_{\tau}[k,j]\Big].
\end{equation}
Minimizing $\mathcal{L}_{\text{SD}}$ discourages skill collapse by pushing different skills to produce dissimilar latent behaviors, complementing $\mathcal{L}_{\text{BSA}}$ and yielding a better-separated skill library.

Finally, we train SkillPlug by minimizing the following weighted sum of objectives:
\begin{equation}
\mathcal{L}
=
\mathcal{L}_{\text{rec}}
+\lambda_{\text{KL}}\mathcal{L}_{\text{KL}}
+\lambda_{\text{BSA}}\mathcal{L}_{\text{BSA}}
+\lambda_{\text{SD}}\mathcal{L}_{\text{SD}}.
\end{equation}

\section{Experiments}
Our experiments aim to answer two questions:
\textit{(1) Can a shared and transferable skill library improve performance when plugged into existing policies, in both multi-task training and few-shot adaptation? (2) What properties make the learned skills transferable across tasks and settings?} Accordingly, we organize our experiments to evaluate overall performance on two simulation benchmarks and a real robot, and to validate these properties empirically.

\subsection{Experimental Setup}
We evaluate SkillPlug on two simulation benchmarks with compact and large-scale backbones. On DISCOVERSE~\cite{jia2025discoverse}, we augment ACT~\cite{zhaolearning}; on LIBERO~\cite{liu2023libero}, we augment OpenVLA-OFT~\cite{kim2502fine}. Across both benchmarks, we use a two-stage protocol: multi-task training to learn a shared skill library, followed by few-shot adaptation that freezes the backbone and skills and fine-tunes only the router and action head. Vanilla baselines follow the same few-shot protocol, except that only the action head is tuned.

\textbf{DISCOVERSE with ACT.}
We train a single policy on 7 tasks with 300 demonstrations per task and no explicit language conditioning. For compact presentation, DISCOVERSE task names are abbreviated by their initials in all tables (e.g., \textit{kiwi place} as K.P.). We use action chunk size 25, train for 6000 epochs with batch size 16 and learning rate $1\times10^{-4}$, and set $K{=}4$. For few-shot adaptation, we start from the multi-task checkpoint and fine-tune a separate policy for each of 4 unseen tasks using 10 demonstrations per task. We follow the standard benchmark evaluation with benchmark-defined reset sampling and 50 trials per task for both multi-task and few-shot settings.

% \textbf{LIBERO with OpenVLA-OFT.} LIBERO comprises $130$ tasks grouped into 4 suites (Spatial, Object, Goal, Long) that probe generalization over layouts, objects, goals, and long-horizon compositions. For multi-task training, we train a separate policy for each suite using all available demonstrations, set $K=8$, and follow the default OpenVLA-OFT training hyperparameters. For few-shot adaptation, we pretrain on LIBERO-Long and adapt to the remaining suites using $5$ demonstrations per task.

\textbf{LIBERO with OpenVLA-OFT.}
LIBERO contains 130 tasks across four suites. We train one policy per suite using all demonstrations, with $K{=}8$ and default OpenVLA-OFT hyperparameters. For few-shot adaptation, we pretrain on LIBERO-Long and adapt to the other suites with 5 demonstrations per task. We follow the original OpenVLA-OFT evaluation protocol with benchmark-defined reset sampling and 500 trials per suite (50 per task).

\subsection{Overall Performance}
\label{subsec:overall_performance}

\textbf{Multi-task performance.}
Table~\ref{tab:discoverse_act_mt} reports the 3-seed mean performance on DISCOVERSE. Vanilla ACT is highly imbalanced across tasks: it performs well on easier pick-and-place tasks such as \textit{kiwi place} ($99.3\!\pm\!0.7$) and \textit{coffeecup place} ($93.3\!\pm\!4.8$), but remains weak on \textit{jujube pick} ($6.7\!\pm\!1.8$), \textit{stack block} ($15.3\!\pm\!1.8$), \textit{block bridge place} ($0.7\!\pm\!0.7$), and \textit{drawer open} ($0.0\!\pm\!0.0$). Adding SkillPlug substantially improves these harder behaviors, raising them to $94.0\!\pm\!3.1$, $80.7\!\pm\!2.4$, $36.7\!\pm\!2.4$, and $100.0\!\pm\!0.0$, respectively, while preserving strong performance on easier tasks. This suggests that the learned skill library expands the behavioral repertoire of a compact ACT backbone. Table~\ref{tab:libero_oft_mt} shows the corresponding LIBERO multi-task results. OpenVLA-OFT is already a strong baseline, achieving $92.9\!\pm\!1.2$, $97.6\!\pm\!0.6$, $96.5\!\pm\!0.9$, and $96.4\!\pm\!0.8$ on LIBERO-Long/Object/Goal/Spatial. SkillPlug remains competitive in this near-saturated regime, improving Long/Object/Goal to $93.5\!\pm\!1.1$, $97.8\!\pm\!0.7$, and $96.7\!\pm\!0.9$, while remaining comparable on Spatial.

\textbf{Few-shot adaptation.}
The same tables also report few-shot adaptation. On DISCOVERSE, with only 10 demonstrations per unseen task, ACT reaches $0.7\!\pm\!0.7$, $36.0\!\pm\!6.9$, $0.0\!\pm\!0.0$, and $82.0\!\pm\!3.1$ on \textit{cuplid cover}, \textit{block place}, \textit{jujube place}, and \textit{mouse push}, respectively. With SkillPlug, these improve to $18.7\!\pm\!6.6$, $56.0\!\pm\!3.1$, $29.3\!\pm\!1.8$, and $87.3\!\pm\!2.4$. Since few-shot adaptation updates only router and action head while keeping the skill library fixed, these gains indicate effective reuse of transferable skills learned during multi-task training. On LIBERO, we pretrain on LIBERO-Long and adapt to the remaining three suites using 5 demonstrations per task. Vanilla OpenVLA-OFT achieves $47.4\!\pm\!2.1$, $45.4\!\pm\!2.2$, and $42.0\!\pm\!2.4$ on LIBERO-Object/Goal/Spatial, whereas SkillPlug improves them to $79.6\!\pm\!1.6$, $83.2\!\pm\!1.7$, and $86.8\!\pm\!2.0$. 

Overall, SkillPlug substantially improves few-shot transfer while preserving strong multi-task performance. This trend holds for both ACT and OpenVLA-OFT, suggesting that its benefit is not tied to a specific backbone scale.

\begin{table*}[t]
\centering
\footnotesize
\caption{DISCOVERSE multi-task and few-shot performance of ACT and ACT+SkillPlug. Results are reported over 3 training seeds.}
\label{tab:discoverse_act_mt}

\begin{tabular*}{\linewidth}{@{\extracolsep{\fill}}l|ccccccc}
\hline
\multicolumn{8}{c}{\textbf{Multi-task}~ $\mathbf{\scriptstyle{(mean~SR(\%) \pm SE)}}$} \\
\hline
Policy & \textit{J.P.} & \textit{K.P.} & \textit{S.B.} & \textit{C.P.} & \textit{B.B.P.} & \textit{D.O.} & \textit{L.C.} \\
\hline
ACT 
& $6.7{ \pm 1.8}$ 
& $\underline{99.3{ \pm 0.7}}$ 
& $15.3{ \pm 1.8}$ 
& $\underline{93.3{ \pm 4.8}}$ 
& $0.7{ \pm 0.7}$ 
& $0.0{ \pm 0.0}$ 
& $72.0{ \pm 4.6}$ \\
+SkillPlug 
& $\mathbf{94.0{ \pm 3.1}}$ 
& $\mathbf{99.3{ \pm 0.7}}$ 
& $\mathbf{80.7{ \pm 2.4}}$ 
& $\mathbf{97.3{ \pm 1.8}}$ 
& $\mathbf{36.7{ \pm 2.4}}$ 
& $\mathbf{100.0{ \pm 0.0}}$ 
& $\mathbf{94.7{ \pm 2.4}}$ \\
\hline
\end{tabular*}

\vspace{1.0em}

\begin{tabular*}{\linewidth}{@{\extracolsep{\fill}}l|cccc}
\hline
\multicolumn{5}{c}{\textbf{Few-shot}~ $\mathbf{\scriptstyle{(mean~SR(\%) \pm SE)}}$}\\
\hline
Policy & \textit{C.C.} & \textit{B.P.} & \textit{J.P.} & \textit{M.P.} \\
\hline
ACT 
& $0.7{ \pm 0.7}$ 
& $36.0{ \pm 6.9}$ 
& $0.0{ \pm 0.0}$ 
& $\underline{82.0{ \pm 3.1}}$ \\
+SkillPlug 
& $\mathbf{18.7{ \pm 6.6}}$ 
& $\mathbf{56.0{ \pm 3.1}}$ 
& $\mathbf{29.3{ \pm 1.8}}$ 
& $\mathbf{87.3{ \pm 2.4}}$ \\
\hline
\end{tabular*}

\end{table*}

\begin{table}[t]
\centering
\footnotesize
\caption{LIBERO multi-task and few-shot performance of OpenVLA-OFT and OpenVLA-OFT+SkillPlug. Results are reported over 3 training seeds.}
\label{tab:libero_oft_mt}

\begin{tabular*}{\columnwidth}{@{\extracolsep{\fill}}l|cccc}
\hline
\multicolumn{5}{c}{\textbf{Multi-task}~ $\mathbf{\scriptstyle{(mean~SR(\%) \pm SE)}}$} \\
\hline
Policy & Long & Object & Goal & Spatial \\
\hline
OpenVLA-OFT 
& $92.9{ \pm 1.2}$ 
& $97.6{ \pm 0.6}$ 
& $96.5{ \pm 0.9}$ 
& $\mathbf{96.4{ \pm 0.8}}$ \\
+SkillPlug 
& $\mathbf{93.5{ \pm 1.1}}$ 
& $\mathbf{97.8{ \pm 0.7}}$ 
& $\mathbf{96.7{ \pm 0.9}}$ 
& $95.6{ \pm 0.7}$ \\
\hline
\end{tabular*}

\vspace{1.0em}

\begin{tabular*}{\columnwidth}{@{\extracolsep{\fill}}l|ccc}
\hline
\multicolumn{4}{c}{\textbf{Few-shot}~ $\mathbf{\scriptstyle{(mean~SR(\%) \pm SE)}}$} \\
\hline
Policy & Object & Goal & Spatial \\
\hline
OpenVLA-OFT 
& $47.4{ \pm 2.1}$ 
& $45.4{ \pm 2.2}$ 
& $42.0{ \pm 2.4}$ \\
+SkillPlug 
& $\mathbf{79.6{ \pm 1.6}}$ 
& $\mathbf{83.2{ \pm 1.7}}$ 
& $\mathbf{86.8{ \pm 2.0}}$ \\
\hline
\end{tabular*}
\end{table}

\subsection{Loss-term ablation} Table~\ref{tab:loss_ablation_compact} reports a progressive loss-term ablation, where each row adds one loss during multi-task pretraining and few-shot results are obtained from the corresponding ablated checkpoint. Ablated variants are single-seed diagnostics; for the full objective, the Avg. column reports the three-seed mean $\pm$ SE. Overall, performance improves as the full objective is added in both settings. KL brings modest gains, raising the average SR from 63.7 to 68.6 in multi-task and from 27.5 to 36.0 in few-shot, suggesting its role as a bottleneck regularizer. SD gives the largest improvement, increasing the averages to 83.7 and 41.0, which indicates the importance of learning distinct, non-redundant skills. BSA further improves the full objective to $86.1{\pm}1.2$ and $47.8{\pm}1.5$, consistent with its role in aligning skills with behavior-level structure for reuse.

\begin{table}[t]
\centering
\footnotesize
\setlength{\tabcolsep}{3.0pt}
\renewcommand{\arraystretch}{1.10}
\caption{Loss-term ablation of ACT+SkillPlug on DISCOVERSE. Ablated variants are single-seed diagnostics; for the full objective, Avg. reports the three-seed mean $\pm$ SE.}
\label{tab:loss_ablation_compact}

\begin{tabular*}{\columnwidth}{@{\extracolsep{\fill}}l|cccccccc}
\hline
\multicolumn{9}{c}{\textbf{Multi-task}~ $\mathbf{\scriptstyle{(SR(\%))}}$} \\
\hline
Loss & \textit{J.P.} & \textit{K.P.} & \textit{S.B.} & \textit{C.P.} & \textit{B.B.P.} & \textit{D.O.} & \textit{L.C.} & \textit{Avg.} \\
\hline
Rec 
& 32.0 & 96.0 & 28.0 & 90.0 & 12.0 & 100.0 & 88.0 & 63.7 \\
+KL 
& 48.0 & 100.0 & 34.0 & 92.0 & 16.0 & 100.0 & 90.0 & 68.6 \\
+SD 
& 88.0 & 100.0 & 78.0 & 100.0 & 24.0 & 100.0 & 96.0 & 83.7 \\
+BSA 
& 100.0 & 100.0 & 82.0 & 100.0 & 38.0 & 100.0 & 98.0 
& $\mathbf{86.1{\scriptstyle \pm 1.2}}$ \\
\hline
\end{tabular*}

\vspace{1.0em}

\begin{tabular*}{\columnwidth}{@{\extracolsep{\fill}}l|ccccc}
\hline
\multicolumn{6}{c}{\textbf{Few-shot}~ $\mathbf{\scriptstyle{(SR(\%))}}$} \\
\hline
Loss & \textit{C.C.} & \textit{B.P.} & \textit{J.P.} & \textit{M.P.} & \textit{Avg.} \\
\hline
Rec 
& 0.0 & 34.0 & 0.0 & 76.0 & 27.5 \\
+KL 
& 2.0 & 50.0 & 10.0 & 82.0 & 36.0 \\
+SD 
& 6.0 & 56.0 & 18.0 & 84.0 & 41.0 \\
+BSA 
& 28.0 & 60.0 & 30.0 & 84.0 
& $\mathbf{47.8{\scriptstyle \pm 1.5}}$ \\
\hline
\end{tabular*}
\end{table}

% \begin{table}[t]
% \centering
% \footnotesize
% \setlength{\tabcolsep}{4.0pt}
% \renewcommand{\arraystretch}{1.10}
% \caption{Loss-term ablation of ACT+SkillPlug on DISCOVERSE.}
% \label{tab:loss_ablation_compact}

% {\bf Multi-task}

% \vspace{0.2em}
% \begin{tabular*}{\columnwidth}{@{\extracolsep{\fill}}lccccccc}
% \hline
% \textbf{Variant} & \textit{JPick} & \textit{KPlace} & \textit{SBlock} & \textit{CPlace} & \textit{BBPlace} & \textit{DOpen} & \textit{LClose} \\
% \hline
% Rec & 32.0 & 96.0 & 28.0 & 90.0 & 12.0 & 100.0 & 88.0 \\
% +KL & 48.0 & 100.0 & 34.0 & 92.0 & 16.0 & 100.0 & 90.0 \\
% +SD & 88.0 & 100.0 & 78.0 & 100.0 & 24.0 & 100.0 & 96.0 \\
% +BSA & \textbf{100.0} & \textbf{100.0} & \textbf{82.0} & \textbf{100.0} & \textbf{38.0} & \textbf{100.0} & \textbf{98.0} \\
% \hline
% \end{tabular*}

% \vspace{0.5em}

% {\bf Few-shot}

% \vspace{0.2em}
% \begin{tabular*}{\columnwidth}{@{\extracolsep{\fill}}lcccc}
% \hline
% \textbf{Variant} & \textit{CCover} & \textit{BPlace} & \textit{JPlace} & \textit{MPush} \\
% \hline
% Rec & 0.0 & 34.0 & 0.0 & 76.0 \\
% +KL & 2.0 & 50.0 & 10.0 & 82.0 \\
% +SD & 6.0 & 56.0 & 18.0 & 84.0 \\
% +BSA & \textbf{28.0} & \textbf{60.0} & \textbf{30.0} & \textbf{84.0} \\
% \hline
% \end{tabular*}
% \end{table}

\subsection{Analysis of Skill Properties}
\label{subsec:skill_properties}

Beyond aggregate success rates, we analyze \emph{what} SkillPlug learns that supports the performance gains observed in multi-task training and few-shot adaptation.
Specifically, we examine the mined skill library from four complementary perspectives:
(1) \emph{scene-agnostic behavior} as evidence of compact, transferable skills;
(2) \emph{compositionality and interpretability} of routing as a mechanism for reusing and recombining skills;
(3) the trade-off between the \emph{number of skills and redundancy} to assess skill separation and redundancy; and
(4) \emph{inference efficiency} as an additional practical benefit at deployment.

\begin{figure}[t]
  \centering
  \includegraphics[width=\linewidth]{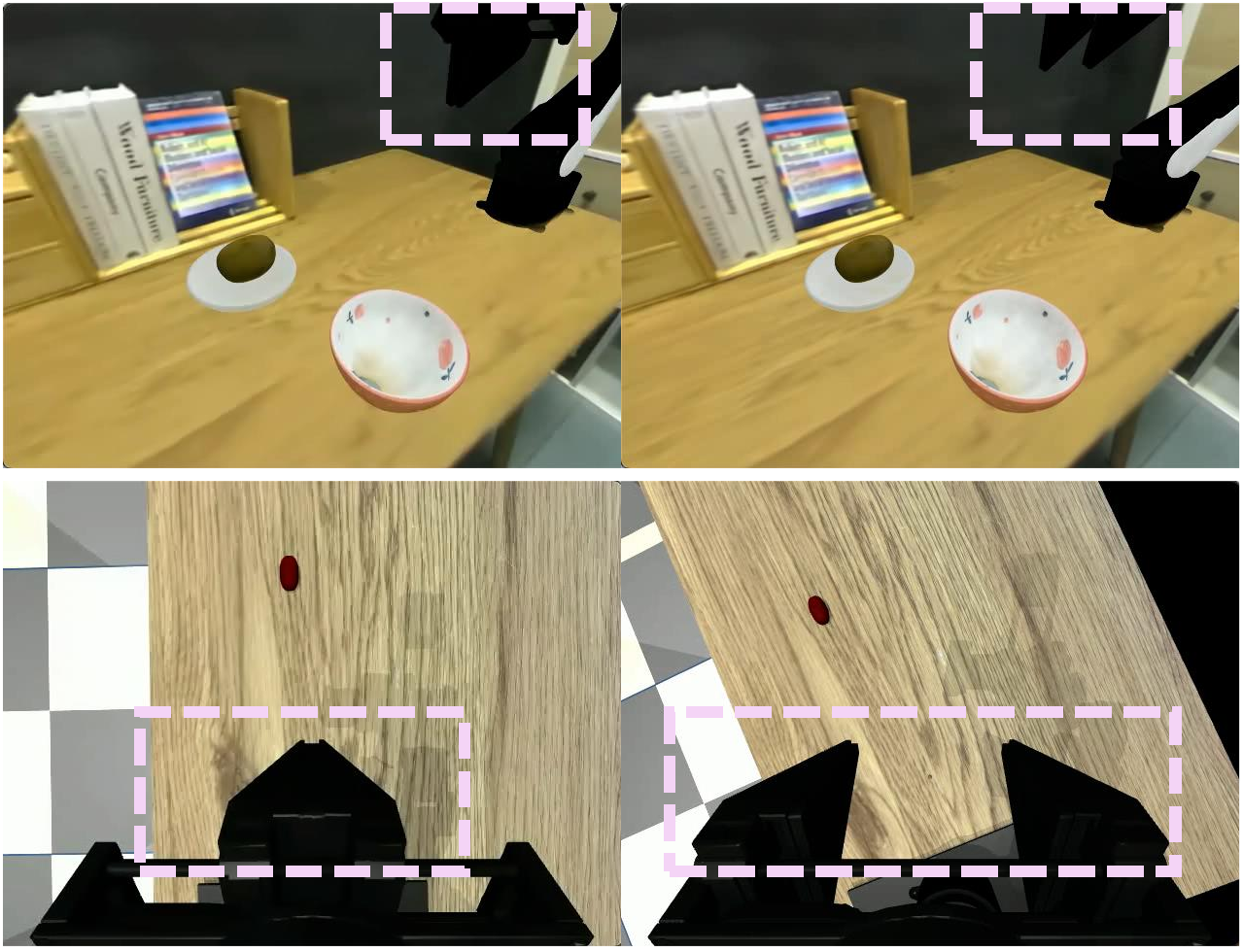}
  \caption{Scene-agnostic behavior of a single skill. The selected skill consistently induces an open–gripper motion. For each scene we show the camera view where this behavior is most clearly visible.}
  \label{fig:scene_agnostic_skill}
\end{figure}

\textbf{Scene-agnostic behavior.} A goal of SkillPlug is to learn scene-agnostic skills that capture reusable behavioral primitives rather than appearance cues, since entanglement with background, texture, or viewpoint can hurt generalization under visual changes. To probe scene-agnosticity, we hard-code the router to always select a fixed skill $k$ and run the resulting policy across tasks with different scene configurations. As shown in Fig.~\ref{fig:scene_agnostic_skill}, the same skill consistently induces an \textit{open-gripper} motion despite substantial appearance changes. This trend holds across skills and tasks, suggesting that SkillPlug captures transferable behavior-level structure that generalizes across both tasks and visual contexts, supporting stronger multi-task coverage and data-efficient few-shot adaptation.

\begin{figure}[t]
  \centering
  \includegraphics[width=\linewidth]{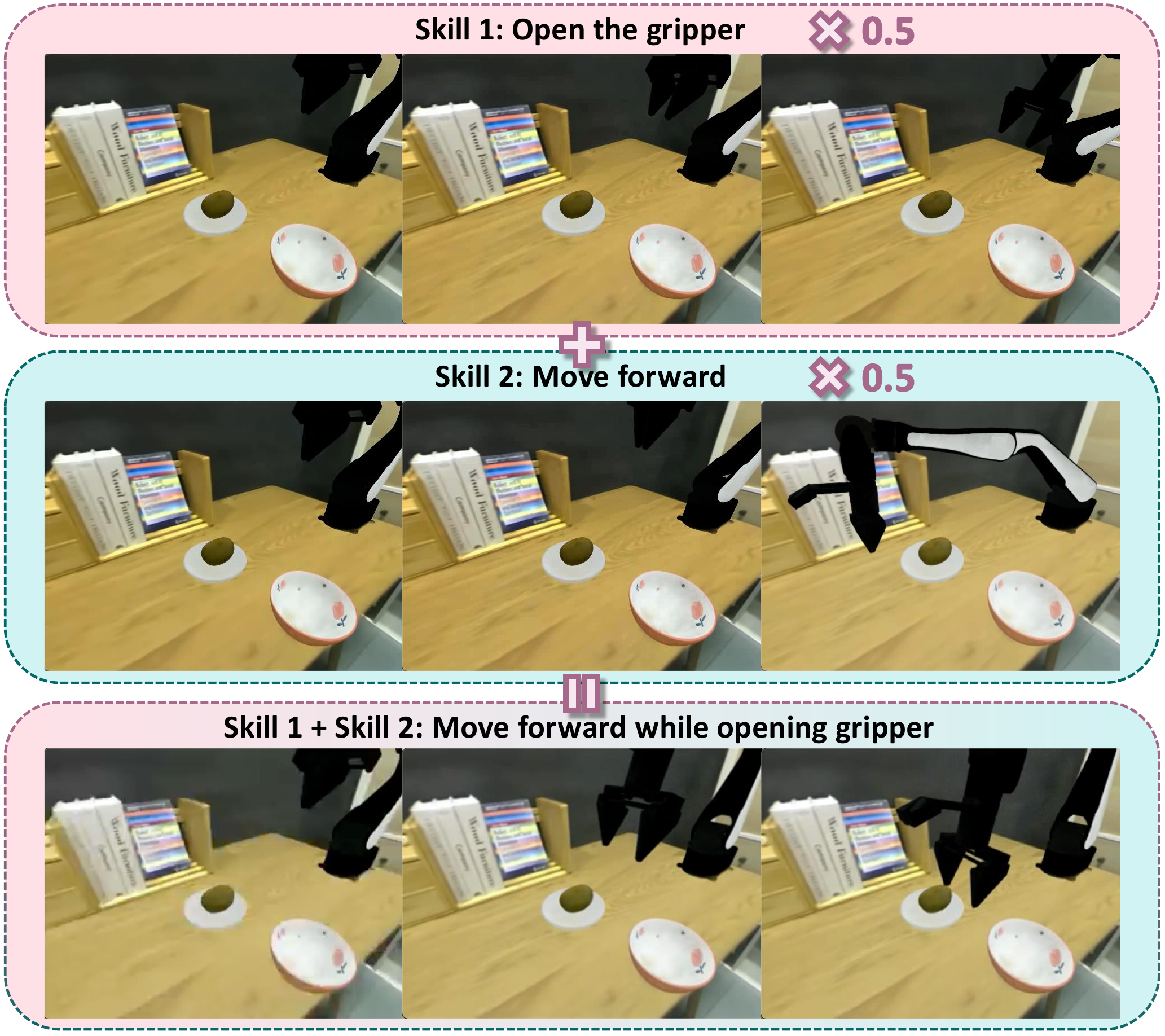}
  \caption{Compositionality via routing. We hard-code the router to select \textit{Skill 1} (top: open gripper), select \textit{Skill 2} (middle: move forward), and assign equal weights to both skills (bottom).  
The mixed routing produces a composed behavior that moves forward while opening the gripper, illustrating that SkillPlug skills can be meaningfully combined by adjusting routing weights.}
  \label{fig:skill_composition}
\end{figure}

\textbf{Compositionality and interpretable routing.}
Beyond scene-agnosticity, we find that the learned skills are compositional and that routing often exhibits coherent structure, supporting adaptation through skill reuse and recombination. To test this, we manually fix the router to mix a pair of learned skills. In Fig.~\ref{fig:skill_composition}, one skill opens the gripper and another moves the end-effector forward. Selecting either skill alone triggers the corresponding primitive, while equal weighting composes them into moving forward while opening the gripper. Similar behavior is observed for other skill pairs. We next examine temporal interpretability. 
% In Fig.~\ref{fig:skill_routing}, we highlight a \textit{kiwi place} episode in which the 3 skills exhibit clear semantics: \textit{Skill 1} opens the gripper, \textit{Skill 2} moves the end-effector right, and \textit{Skill 3} moves it left. From $t{=}0$ to $t{=}40$, \textit{Skill~1} is dominant and \textit{Skill~2} gradually increases as the robot approaches the kiwi with an open gripper. Around $t{\approx}40$ to $t{\approx}60$, \textit{Skill~2} drops and the gripper closes to grasp the kiwi. After $t{=}60$, \textit{Skill~3} becomes dominant as the robot transports the kiwi toward the bowl. \textcolor{red}{Overall, these visualizations suggest phase-structured activations over reusable motion primitives; although we do not impose explicit objective for routing interpretability, this pattern emerges in practice as the router schedules a well-structured skill library to minimize action reconstruction.}

\begin{figure}[t]
  \centering
  \includegraphics[width=\linewidth]{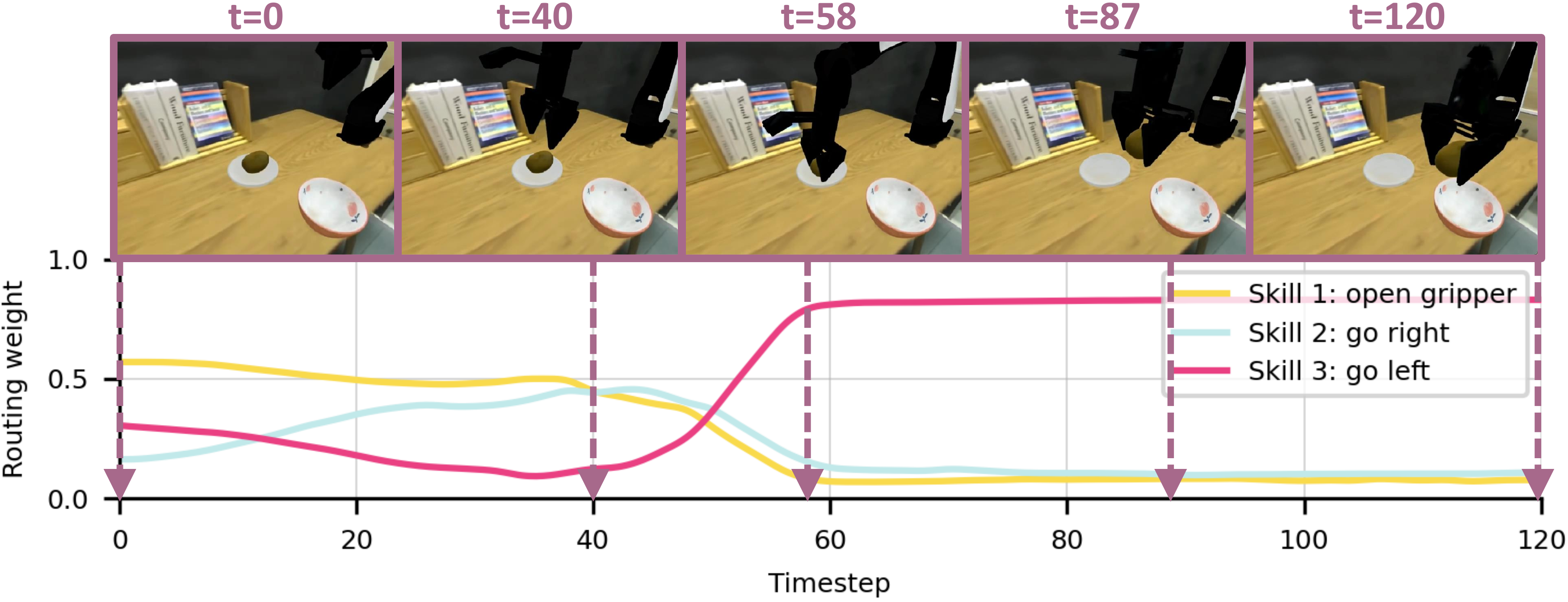}
  \caption{Interpretable routing dynamics. Top: key frames of a \textit{kiwi place} episode. Bottom: routing weights for three skills. \textit{Skill~1} dominates during approach with an open gripper ($t{=}0$--$40$) while \textit{Skill~2} ramps up; \textit{Skill~2} then drops as the gripper closes to grasp the kiwi ($t{\approx}40$--$60$); \textit{Skill~3} becomes dominant during transport to the bowl ($t{>}60$).}
  \label{fig:skill_routing}
\end{figure}

In Fig.~\ref{fig:skill_routing}, a \textit{kiwi place} episode shows clear semantics: \textit{Skill~1} opens the gripper, \textit{Skill~2} moves the end-effector right, and \textit{Skill~3} moves it left. The routing is phase-structured: \textit{Skill~1} and \textit{Skill~2} dominate during approach, \textit{Skill~2} drops during grasping, and \textit{Skill~3} becomes dominant during transport to the bowl. Overall, these visualizations suggest phase-structured activations over reusable motion primitives; although we do not impose explicit objective for routing interpretability, this pattern emerges in practice as the router schedules a well-structured skill library to minimize action reconstruction.

% \textbf{Redundancy vs.\ number of skills.}
% We study how the number of skills $K$ affects performance and redundancy. Table~\ref{tab:num_skills_quant} reports DISCOVERSE success for $K\in\{2,4,6\}$. With $K{=}2$, ACT+SkillPlug already performs well on average (76\%) but struggles on harder tasks such as \textit{stack block} and \textit{block bridge place}. Increasing to $K{=}4$ substantially improves the average to 88\%, with gains of +12\% and +36\% on these two tasks, while $K{=}6$ yields only marginal improvement (90\%). Overall, most gains are achieved at $K{=}4$, suggesting a favorable expressivity--cost trade-off.

% To explain the saturation, we analyze pairwise cosine similarities between skill embeddings. Fig.~\ref{fig:cos_sim}(a) shows near-zero off-diagonal similarity for $K{=}2$, indicating well-separated skills; mild correlations appear at $K{=}4$; and redundancy increases at $K{=}6$ as multiple skills become more aligned. This is consistent with routing dynamics in Fig.~\ref{fig:cos_sim}(b), which shows an episode with $K{=}4$ where \textit{Skill~1} and \textit{Skill~2} follow similar activations over time, matching their higher embedding similarity.

% Notably, although we have 7 training tasks, redundancy already emerges at $K{=}6$, indicating that learned skills are reusable primitives rather than one-to-one surrogates of task labels, and that overly large $K$ introduces redundant variants.

\begin{figure}[t]
  \centering
  \includegraphics[width=\linewidth]{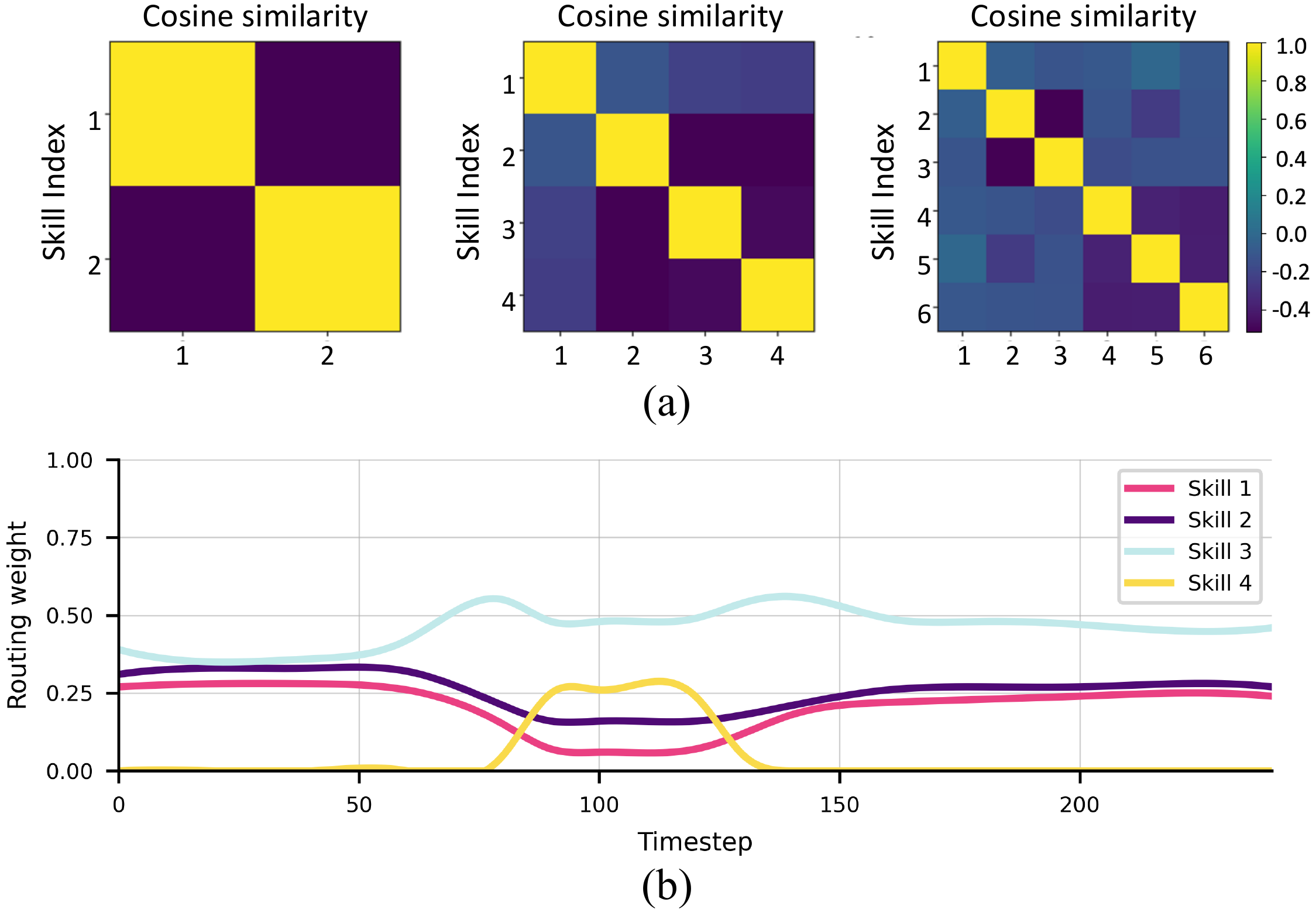}
    \caption{Skill diversity and routing dynamics.
        (a) Cosine-similarity matrices between skill embeddings for different numbers of skills $K$ (left: $K{=}2$, middle: $K{=}4$, right: $K{=}6$); off-diagonal values reflect redundancy, with smaller values indicating more diverse skills.
        (b) Routing weights for a episode with $K{=}4$, where \emph{Skill 1} and \emph{Skill 2} follow very similar activation patterns, consistent with their high off-diagonal similarity in (a), revealing redundancy between them, while the other skills remain more distinct.}
  \label{fig:cos_sim}
\end{figure}

\textbf{Redundancy vs.\ number of skills.}
We study how the number of skills $K$ affects performance and redundancy. Table~\ref{tab:num_skills_quant} reports DISCOVERSE success for $K\in\{2,4,6\}$. Performance improves from 76\% at $K{=}2$ to 88\% at $K{=}4$, with notable gains on harder tasks such as \textit{stack block} (+12\%) and \textit{block bridge place} (+36\%), while increasing to $K{=}6$ brings only a marginal gain to 90\%. Thus, most benefits are achieved at $K{=}4$. To explain this saturation, we analyze pairwise cosine similarities between skill embeddings. Fig.~\ref{fig:cos_sim}(a) shows that skills are well separated at $K{=}2$, mild redundancy appears at $K{=}4$, and redundancy increases at $K{=}6$. Fig.~\ref{fig:cos_sim}(b) is consistent with this trend: for $K{=}4$, \textit{Skill~1} and \textit{Skill~2} exhibit similar temporal activations, matching their higher embedding similarity. Notably, although the training suite contains 7 tasks, redundancy already emerges at $K{=}6$, suggesting that learned skills are reusable primitives rather than one-to-one surrogates of task labels, and that overly large $K$ mainly introduces redundant variants.

\begin{table}[t]
\centering
\footnotesize
% \caption{DISCOVERSE per-task success rates of ACT+SkillPlug for different numbers of skills $K\in\{2,4,6\}$.}
\caption{DISCOVERSE per-task success rates of ACT+SkillPlug for different numbers of skills $K\in\{2,4,6\}$ (single seed).}
\label{tab:num_skills_quant}
\begin{tabular*}{\columnwidth}{@{\extracolsep{\fill}}l|ccccccc}
\hline
$K$ & \textit{J.P.} & \textit{K.P.} & \textit{S.B.} & \textit{C.P.} & \textit{B.B.P.} & \textit{D.O.} & \textit{L.C.} \\
\hline
2 & 98.0 & 100.0 & 70.0 & 70.0 & 2.0 & 100.0 & 92.0 \\
4 & \underline{100.0} & \underline{100.0} & 82.0 & \underline{100.0} & 38.0 & \underline{100.0} & 98.0 \\
6 & \textbf{100.0} & \textbf{100.0} & \textbf{86.0} & \textbf{100.0} & \textbf{46.0} & \textbf{100.0} & \textbf{100.0} \\
\hline
\end{tabular*}
\end{table}

% \begin{table}[t]
% \centering
% \footnotesize
% \caption{DISCOVERSE per-task success rates of ACT+SkillPlug for different numbers of skills $K\in\{2,4,6\}$.}
% \label{tab:num_skills_quant}
% \begin{tabular*}{\columnwidth}{@{\extracolsep{\fill}}l|ccc}
% % \begin{tabular}{l|ccc}
% \hline
% Task & 2 skills & 4 skills & 6 skills \\
% \hline
% \textit{jujube pick}         & 98.0 & \underline{100.0} & \textbf{100.0} \\
% \textit{kiwi place}          & 100.0 & \underline{100.0} & \textbf{100.0} \\
% \textit{stack block}         & 70.0 & 82.0            & \textbf{86.0} \\
% \textit{coffeecup place}     & 70.0 & \underline{100.0} & \textbf{100.0} \\
% \textit{block bridge place}  & 2.0 & 38.0             & \textbf{46.0} \\
% \textit{drawer open}         & 100.0 & \underline{100.0} & \textbf{100.0} \\
% \textit{laptop close}        & 92.0 & 98.0 & \textbf{100.0} \\
% \hline
% \end{tabular*}
% \end{table}

% \textbf{Inference efficiency.}
% Finally, we empirically observe that the learned skills also improve inference efficiency. For ACT-based policies on DISCOVERSE, we keep the action chunk size fixed and measure the number of model inference steps per successful episode. Table~\ref{tab:inference_steps} reports this quantity for the multi-task ACT baseline and ACT+SkillPlug.
% ACT+SkillPlug requires fewer inference steps on every task, with especially large reductions on harder tasks such as \textit{stack block} (126$\rightarrow$86) and \textit{block bridge place} (181$\rightarrow$119).
% These efficiency gains come on top of the success-rate improvements in Sec.~\ref{subsec:overall_performance}.
% We attribute this to temporal abstraction in the mined skills.
% By providing a stronger prior over how actions evolve within an action chunk, SkillPlug produces more coherent control per policy query, allowing the agent to reach the goal with fewer model evaluations while improving success rates.

\textbf{Inference efficiency.}
Finally, we observe that the learned skills also improve inference efficiency. For ACT-based policies on DISCOVERSE, we fix the action chunk size and measure the number of model inference steps per successful episode. As shown in Table~\ref{tab:inference_steps}, ACT+SkillPlug requires fewer steps on every task, with especially large reductions on harder tasks such as \textit{stack block} (126$\rightarrow$86) and \textit{block bridge place} (181$\rightarrow$119). These gains come in addition to the success-rate improvements in Sec.~\ref{subsec:overall_performance}, suggesting that the learned skills provide a stronger temporal prior that enables more efficient control.

% \begin{table}[t]
% \centering
% \footnotesize
% \caption{Average number of inference steps per successful episode on DISCOVERSE.
% Lower is better.}
% \label{tab:inference_steps}
% \begin{tabular*}{\columnwidth}{@{\extracolsep{\fill}}l|cc}
% % \begin{tabular}{l|cc}
% \hline
% Task & ACT & ACT+SkillPlug \\
% \hline
% \textit{jujube pick}         & 79  & \textbf{48} \\
% \textit{kiwi place}          & 140 & \textbf{96} \\
% \textit{stack block}         & 126 & \textbf{86} \\
% \textit{coffeecup place}     & 148 & \textbf{116} \\
% \textit{block bridge place}  & 181 & \textbf{119} \\
% \textit{drawer open}         & 72  & \textbf{62} \\
% \textit{laptop close}        & 96  & \textbf{84} \\
% \hline
% \end{tabular*}
% \end{table}

\begin{table}[t]
\centering
\footnotesize
\caption{Average number of inference steps per successful episode on DISCOVERSE.
Lower is better.}
\label{tab:inference_steps}
\begin{tabular*}{\columnwidth}{@{\extracolsep{\fill}}l|ccccccc}
\hline
Policy & \textit{J.P.} & \textit{K.P.} & \textit{S.B.} & \textit{C.P.} & \textit{B.B.P.} & \textit{D.O.} & \textit{L.C.} \\
% Method & \textit{jujube pick} & \textit{kiwi place} & \textit{stack block} & \textit{coffeecup place} & \textit{block bridge place} & \textit{drawer open} & \textit{laptop close} \\
\hline
ACT & 79 & 140 & 126 & 148 & 181 & 72 & 96 \\
+SkillPlug & \textbf{48} & \textbf{96} & \textbf{86} & \textbf{116} & \textbf{119} & \textbf{62} & \textbf{84} \\
\hline
\end{tabular*}
\end{table}

\subsection{Real-World Performance}

\textbf{Experimental setup.}
We further evaluate ACT+SkillPlug on a real-world tabletop manipulation platform with a \emph{Galaxea A1} 6-DoF arm, a parallel gripper, a wrist-mounted RealSense D435i, and an external RealSense L515.
We follow the same train-then-adapt protocol as in simulation: multi-task training uses 6 contact-rich tasks, and few-shot uses 3 unseen tasks with only 5 demonstrations each.

\begin{figure}[t]
  \centering
  \includegraphics[width=\linewidth]{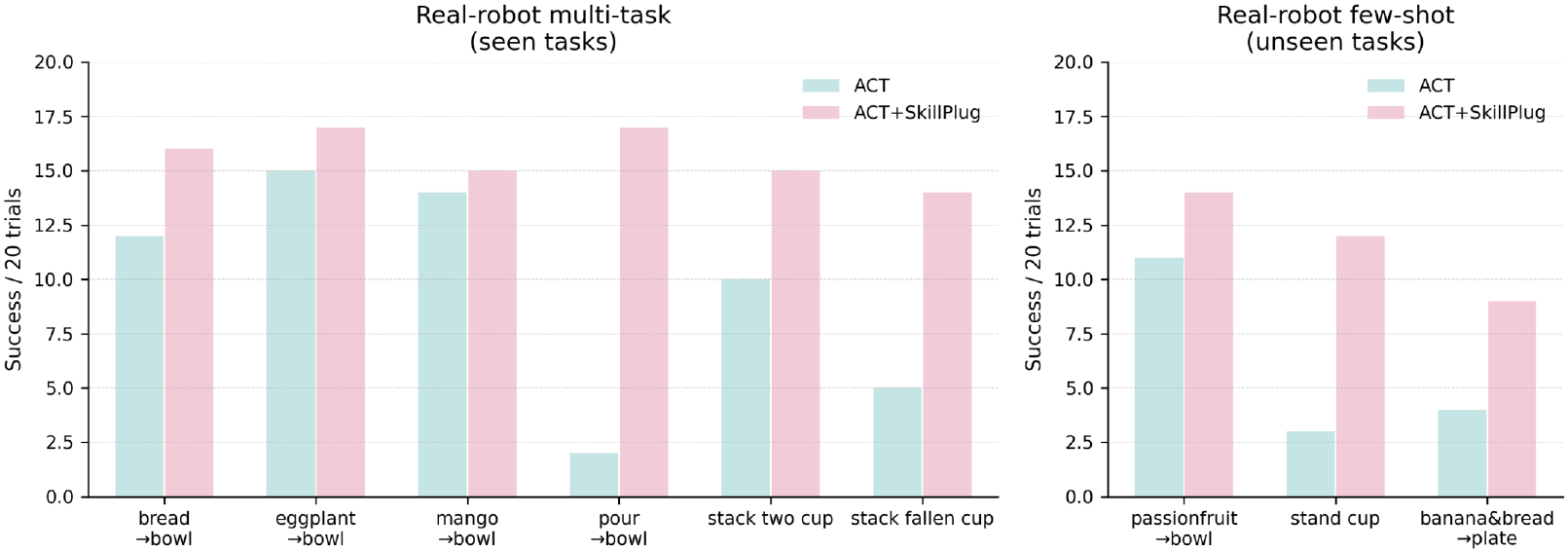}
  \caption{Real-robot performance of ACT and ACT+SkillPlug. Left: multi-task training on 6 seen tasks. Right: few-shot fine-tuning on 3 unseen tasks with 5 demonstrations per task. Bars show the number of successes out of 20 trials; ACT+SkillPlug consistently improves over ACT on all tasks.}
  \label{fig:real_exp}
\end{figure}

\begin{figure*}[t]
  \centering
  \resizebox{\linewidth}{0.28\height}{%
\includegraphics{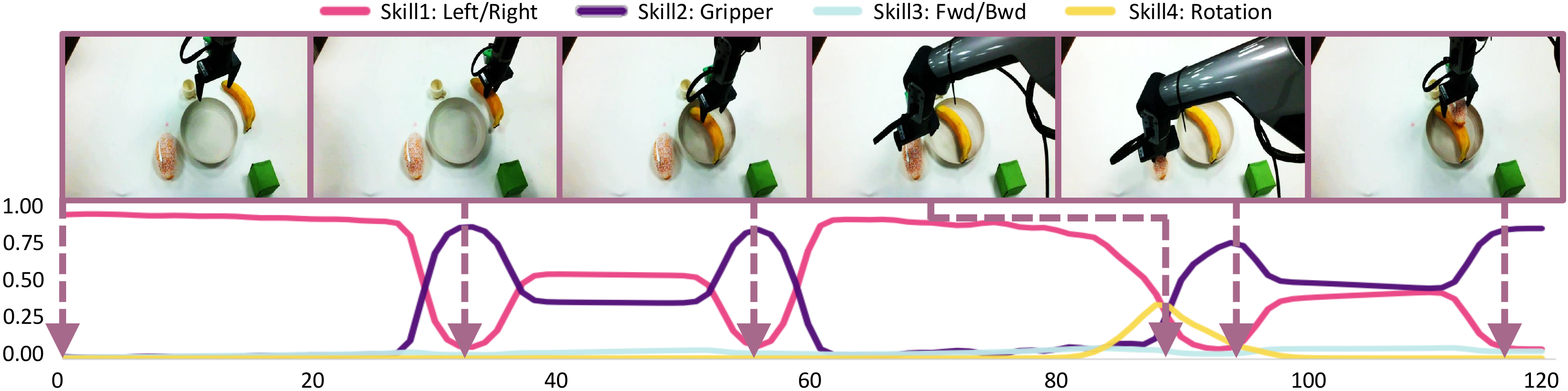}%
  }
  \caption{Skill routing visualization on a real robot for \textit{pick banana and bread on the white plate} task in the few-shot setting. The router exhibits two similar pick-and-place phases: motion-dominant segments uses Skill 1, gripper state changes uses Skill 2, and a Skill 4 is triggered near $~90$ steps reflects wrist rotation to align the end-effector for grasping the bread.}
  \label{fig:realrobot_routing}
\end{figure*}

\textbf{Performance Analysis.} Fig.~\ref{fig:real_exp} shows that ACT+SkillPlug consistently improves both real-robot multi-task training and few-shot adaptation.
On the 6 seen tasks, it improves the average success count from 9.7/20 to 15.7/20 per task.
On the 3 unseen tasks, it improves the few-shot average from 6/20 to 11.7/20, with especially large gains on \emph{stand cup} (3$\rightarrow$12) and \emph{banana \& bread $\rightarrow$ plate} (4$\rightarrow$9). Furthermore, Fig.~\ref{fig:realrobot_routing} visualizes routing weights for a real-robot episode of \textit{pick banana and bread on the white plate}.
The routing shows two similar segments corresponding to the two pick-and-place phases.
In both, \textit{Skill~1} dominates during arm motion and \textit{Skill~2} around gripper state changes, while \textit{Skill~4} briefly activates near $\sim$90 steps, likely capturing wrist rotation for grasping the bread.

% \begin{figure*}[t]
%   \centering
%   \includegraphics[width=\linewidth]{figures/realrobot_routing_v2.pdf}
%   \caption{Skill routing visualization on a real robot for \textit{pick banana and bread on the white plate} task in the few-shot setting. The router exhibits two similar pick-and-place phases: motion-dominant segments uses Skill 1, gripper state changes uses Skill 2, and a Skill 4 is triggered near $~90$ steps reflects wrist rotation to align the end-effector for grasping the bread.}
%   \label{fig:realrobot_routing}
% \end{figure*}

\section{Conclusions}

We presented \emph{SkillPlug}, a plug-in, architecture-agnostic framework for unsupervised skill mining that augments existing visuomotor policies with a lightweight skill-conditioning module to learn a \emph{transferable} and \emph{reusable} skill library from raw multi-task demonstrations. These transferable skills serve as a shared prior that improves both multi-task performance and few-shot adaptation. One limitation is that the library remains fixed after multi-task training, which may hinder adaptation when novel tasks require new primitives. An important future direction is to make the library expandable in continual and cross-embodiment settings.

\bibliographystyle{IEEEtran}
\bibliography{refs}

@article{jia2025discoverse,
  title={DISCOVERSE: Efficient robot simulation in complex high-fidelity environments},
  author={Jia, Yufei and Wang, Guangyu and Dong, Yuhang and Wu, Junzhe and Zeng, Yupei and Lin, Haonan and Wang, Zifan and Ge, Haizhou and Gu, Weibin and Ding, Kairui and others},
  journal={arXiv preprint arXiv:2507.21981},
  year={2025}
}

@inproceedings{zhaolearning,
  title={Learning Fine-Grained Bimanual Manipulation with Low-Cost Hardware},
  author={Zhao, Tony Z and Kumar, Vikash and Levine, Sergey and Finn, Chelsea},
  booktitle={ICML Workshop on New Frontiers in Learning, Control, and Dynamical Systems}
}

@article{kim2502fine,
  title={Fine-tuning vision-language-action models: Optimizing speed and success, 2025},
  author={Kim, Moo Jin and Finn, Chelsea and Liang, Percy},
  journal={URL https://arxiv. org/abs/2502.19645}
}

@article{liu2023libero,
  title={Libero: Benchmarking knowledge transfer for lifelong robot learning},
  author={Liu, Bo and Zhu, Yifeng and Gao, Chongkai and Feng, Yihao and Liu, Qiang and Zhu, Yuke and Stone, Peter},
  journal={Advances in Neural Information Processing Systems},
  volume={36},
  pages={44776--44791},
  year={2023}
}

@article{xu2025query,
  title={Query-Centric Diffusion Policy for Generalizable Robotic Assembly},
  author={Xu, Ziyi and Lin, Haohong and Liu, Shiqi and Zhao, Ding},
  journal={arXiv preprint arXiv:2509.18686},
  year={2025}
}

@inproceedings{lorang2025few,
  title={Few-Shot Neuro-Symbolic Imitation Learning for Long-Horizon Planning and Acting},
  author={Lorang, Pierrick and Lu, Hong and Huemer, Johannes and Zips, Patrik and Scheutz, Matthias},
  booktitle={Conference on Robot Learning},
  pages={2501--2518},
  year={2025},
  organization={PMLR}
}

@article{zhao2025generalizable,
  title={Generalizable Hierarchical Skill Learning via Object-Centric Representation},
  author={Zhao, Haibo and Qi, Yu and Hu, Boce and Zhu, Yizhe and Chen, Ziyan and Tian, Heng and Zhu, Xupeng and Howell, Owen and Huang, Haojie and Walters, Robin and others},
  journal={arXiv preprint arXiv:2510.21121},
  year={2025}
}

@inproceedings{fan2025long,
  title={Long-VLA: Unleashing Long-Horizon Capability of Vision Language Action Model for Robot Manipulation},
  author={Fan, Yiguo and Bai, Shuanghao and Tong, Xinyang and Ding, Pengxiang and Zhu, Yuyang and Lu, Hongchao and Dai, Fengqi and Zhao, Wei and Liu, Yang and Huang, Siteng and others},
  booktitle={Conference on Robot Learning},
  pages={2018--2037},
  year={2025},
  organization={PMLR}
}

@article{wang2025experts,
  title={From experts to a generalist: Toward general whole-body control for humanoid robots},
  author={Wang, Yuxuan and Yang, Ming and Ding, Ziluo and Zhang, Yu and Zeng, Weishuai and Xu, Xinrun and Jiang, Haobin and Lu, Zongqing},
  journal={arXiv preprint arXiv:2506.12779},
  year={2025}
}

@inproceedings{zheng2025universal,
  title={Universal actions for enhanced embodied foundation models},
  author={Zheng, Jinliang and Li, Jianxiong and Liu, Dongxiu and Zheng, Yinan and Wang, Zhihao and Ou, Zhonghong and Liu, Yu and Liu, Jingjing and Zhang, Ya-Qin and Zhan, Xianyuan},
  booktitle={Proceedings of the Computer Vision and Pattern Recognition Conference},
  pages={22508--22519},
  year={2025}
}

@article{cheng2025moe,
  title={MoE-DP: An MoE-Enhanced Diffusion Policy for Robust Long-Horizon Robotic Manipulation with Skill Decomposition and Failure Recovery},
  author={Cheng, Baiye and Liang, Tianhai and Huang, Suning and Shao, Maanping and Zhang, Feihong and Xu, Botian and Xue, Zhengrong and Xu, Huazhe},
  journal={arXiv preprint arXiv:2511.05007},
  year={2025}
}

@inproceedings{fu2024language,
  title={Language-guided skill learning with temporal variational inference},
  author={Fu, Haotian and Sharma, Pratyusha and Stengel-Eskin, Elias and Konidaris, George and Le Roux, Nicolas and C{\^o}t{\'e}, Marc-Alexandre and Yuan, Xingdi},
  booktitle={Proceedings of the 41st International Conference on Machine Learning},
  pages={14135--14156},
  year={2024}
}

@inproceedings{jiang2025discrete,
  title={Discrete Latent Plans via Semantic Skill Abstractions.},
  author={Jiang, Haobin and Wang, Jiangxing and Lu, Zongqing},
  booktitle={ICLR},
  year={2025}
}

@inproceedings{liusemantic,
  title={Semantic Temporal Abstraction via Vision-Language Model Guidance for Efficient Reinforcement Learning},
  author={Liu, Tian-Shuo and Liu, Xu-Hui and Chen, Ruifeng and Jin, Lixuan and Wang, Pengyuan and Zhang, Zhilong and Yu, Yang},
  booktitle={The Thirteenth International Conference on Learning Representations}
}

@article{qi2025compose,
  title={Compose by Focus: Scene Graph-based Atomic Skills},
  author={Qi, Han and Chen, Changhe and Yang, Heng},
  journal={arXiv preprint arXiv:2509.16053},
  year={2025}
}

@inproceedings{yao2025think,
  title={Think Small, Act Big: Primitive Prompt Learning for Lifelong Robot Manipulation},
  author={Yao, Yuanqi and Liu, Siao and Song, Haoming and Qu, Delin and Chen, Qizhi and Ding, Yan and Zhao, Bin and Wang, Zhigang and Li, Xuelong and Wang, Dong},
  booktitle={Proceedings of the Computer Vision and Pattern Recognition Conference},
  pages={22573--22583},
  year={2025}
}

@article{mete2024quest,
  title={Quest: Self-supervised skill abstractions for learning continuous control},
  author={Mete, Atharva and Xue, Haotian and Wilcox, Albert and Chen, Yongxin and Garg, Animesh},
  journal={Advances in Neural Information Processing Systems},
  volume={37},
  pages={4062--4089},
  year={2024}
}

@article{xu2025speci,
  title={SPECI: Skill Prompts based Hierarchical Continual Imitation Learning for Robot Manipulation},
  author={Xu, Jingkai and Nie, Xiangli},
  journal={arXiv preprint arXiv:2504.15561},
  year={2025}
}

@article{chi2025diffusion,
  title={Diffusion policy: Visuomotor policy learning via action diffusion},
  author={Chi, Cheng and Xu, Zhenjia and Feng, Siyuan and Cousineau, Eric and Du, Yilun and Burchfiel, Benjamin and Tedrake, Russ and Song, Shuran},
  journal={The International Journal of Robotics Research},
  volume={44},
  number={10-11},
  pages={1684--1704},
  year={2025},
  publisher={Sage Publications Sage UK: London, England}
}

@inproceedings{mandi2022towards,
  title={Towards more generalizable one-shot visual imitation learning},
  author={Mandi, Zhao and Liu, Fangchen and Lee, Kimin and Abbeel, Pieter},
  booktitle={2022 International conference on robotics and automation (ICRA)},
  pages={2434--2444},
  year={2022},
  organization={IEEE}
}

@article{brohan2022rt,
  title={Rt-1: Robotics transformer for real-world control at scale},
  author={Brohan, Anthony and Brown, Noah and Carbajal, Justice and Chebotar, Yevgen and Dabis, Joseph and Finn, Chelsea and Gopalakrishnan, Keerthana and Hausman, Karol and Herzog, Alex and Hsu, Jasmine and others},
  journal={arXiv preprint arXiv:2212.06817},
  year={2022}
}

@inproceedings{zitkovich2023rt,
  title={Rt-2: Vision-language-action models transfer web knowledge to robotic control},
  author={Zitkovich, Brianna and Yu, Tianhe and Xu, Sichun and Xu, Peng and Xiao, Ted and Xia, Fei and Wu, Jialin and Wohlhart, Paul and Welker, Stefan and Wahid, Ayzaan and others},
  booktitle={Conference on Robot Learning},
  pages={2165--2183},
  year={2023},
  organization={PMLR}
}

@inproceedings{kim2025openvla,
  title={OpenVLA: An Open-Source Vision-Language-Action Model},
  author={Kim, Moo Jin and Pertsch, Karl and Karamcheti, Siddharth and Xiao, Ted and Balakrishna, Ashwin and Nair, Suraj and Rafailov, Rafael and Foster, Ethan P and Sanketi, Pannag R and Vuong, Quan and others},
  booktitle={Conference on Robot Learning},
  pages={2679--2713},
  year={2025},
  organization={PMLR}
}

@inproceedings{mees2024octo,
  title={Octo: An open-source generalist robot policy},
  author={Mees, Oier and Ghosh, Dibya and Pertsch, Karl and Black, Kevin and Walke, Homer Rich and Dasari, Sudeep and Hejna, Joey and Kreiman, Tobias and Xu, Charles and Luo, Jianlan and others},
  booktitle={First Workshop on Vision-Language Models for Navigation and Manipulation at ICRA 2024},
  year={2024}
}

@article{black2024pi,
  title={$\pi$: A Vision-Language-Action Flow Model for General Robot Control},
  author={Black, Kevin and Brown, Noah and Driess, Danny and Esmail, Adnan and Equi, Michael and Finn, Chelsea and Fusai, Niccolo and Groom, Lachy and Hausman, Karol and Ichter, Brian and others},
  journal={CoRR},
  year={2024}
}

@article{intelligence2504pi0,
  title={$\pi$0. 5: A vision-language-action model with open-world generalization. arXiv 2025},
  author={Intelligence, P and Black, K and Brown, N and Darpinian, J and Dhabalia, K and Driess, D and Esmail, A and Equi, M and Finn, C and Fusai, N and others},
  journal={arXiv preprint arXiv:2504.16054}
}

@inproceedings{liurdt,
  title={RDT-1B: a Diffusion Foundation Model for Bimanual Manipulation},
  author={Liu, Songming and Wu, Lingxuan and Li, Bangguo and Tan, Hengkai and Chen, Huayu and Wang, Zhengyi and Xu, Ke and Su, Hang and Zhu, Jun},
  booktitle={The Thirteenth International Conference on Learning Representations}
}

@article{touvron2023llama,
  title={Llama: Open and efficient foundation language models},
  author={Touvron, Hugo and Lavril, Thibaut and Izacard, Gautier and Martinet, Xavier and Lachaux, Marie-Anne and Lacroix, Timoth{\'e}e and Rozi{\`e}re, Baptiste and Goyal, Naman and Hambro, Eric and Azhar, Faisal and others},
  journal={arXiv preprint arXiv:2302.13971},
  year={2023}
}

@article{bai2023qwenvl,
  title={Qwen-VL: A Versatile Vision-Language Model for Understanding, Localization, Text Reading, and Beyond},
  author={Bai, Jinze and Bai, Shuai and Yang, Shusheng and Wang, Shijie and Tan, Sinan and Wang, Peng and Lin, Junyang and Zhou, Chang and Zhou, Jingren},
  journal={arXiv preprint arXiv:2308.12966},
  year={2023}
}

@article{beyer2024paligemma,
  title={PaliGemma: A versatile 3B VLM for transfer},
  author={Beyer, Lucas and Steiner, Andreas and Pinto, Andr{\'e} Susano and Kolesnikov, Alexander and Wang, Xiao and Salz, Daniel and Neumann, Maxim and Alabdulmohsin, Ibrahim and Tschannen, Michael and Bugliarello, Emanuele and others},
  journal={CoRR},
  year={2024}
}

@inproceedings{walke2023bridgedata,
  title={Bridgedata v2: A dataset for robot learning at scale},
  author={Walke, Homer Rich and Black, Kevin and Zhao, Tony Z and Vuong, Quan and Zheng, Chongyi and Hansen-Estruch, Philippe and He, Andre Wang and Myers, Vivek and Kim, Moo Jin and Du, Max and others},
  booktitle={Conference on Robot Learning},
  pages={1723--1736},
  year={2023},
  organization={PMLR}
}

@inproceedings{o2024open,
  title={Open X-Embodiment: Robotic Learning Datasets and RT-X Models: Open X-Embodiment Collaboration},
  author={O'Neill, Abby and Rehman, Abdul and Maddukuri, Abhiram and Gupta, Abhishek and Padalkar, Abhishek and Lee, Abraham and Pooley, Acorn and Gupta, Agrim and Mandlekar, Ajay and Jain, Ajinkya and others},
  booktitle={ICRA},
  year={2024}
}

@article{li2025cronusvla,
  title={CronusVLA: Transferring Latent Motion Across Time for Multi-Frame Prediction in Manipulation},
  author={Li, Hao and Yang, Shuai and Chen, Yilun and Tian, Yang and Yang, Xiaoda and Chen, Xinyi and Wang, Hanqing and Wang, Tai and Zhao, Feng and Lin, Dahua and others},
  journal={arXiv preprint arXiv:2506.19816},
  year={2025}
}

\vfill

\end{document}